%% file: egpaper_for_review.tex
\documentclass[10pt,twocolumn,letterpaper]{article}
\usepackage[accsupp]{axessibility}  
\usepackage{iccv}
\usepackage{times}
\usepackage{epsfig}
\usepackage{graphicx}
\usepackage{amsmath}
\usepackage{amssymb}
\usepackage{float}
\usepackage{caption}
\usepackage{lipsum}

\usepackage{enumitem}
\usepackage{bm}
\usepackage{booktabs}
\usepackage{multirow}

\usepackage[pagebackref=true,breaklinks=true,letterpaper=true,colorlinks,bookmarks=false]{hyperref}

\iccvfinalcopy 

\def\iccvPaperID{4175} 

\ificcvfinal\fi

\begin{document}

\title{Implicit Identity Representation Conditioned Memory Compensation Network for Talking Head Video Generation}


\author{Fa-Ting Hong\\
CSE, HKUST\\
{\tt\small fhongac@connect.ust.hk}
\and
Dan Xu\thanks{Corresponding author}\\
CSE, HKUST\\
{\tt\small danxu@cse.ust.hk}
}
\twocolumn[{
\maketitle
\begin{center}
    \captionsetup{type=figure}
    \includegraphics[width=1\textwidth]{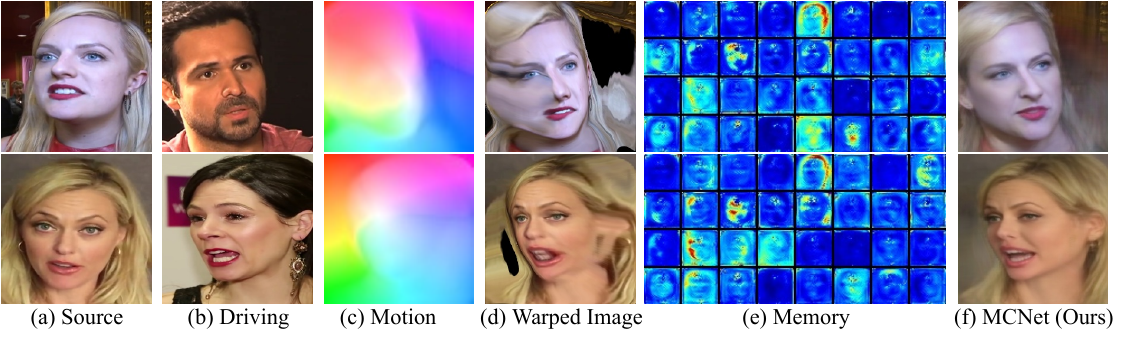}
    \captionof{figure}{The animation illustration of the proposed implicit identity representation conditioned memory compensation network (MCNet). MCNet first learns the motion flow (\textbf{c}) between the source and driving images; (\textbf{d}) shows possible occlusion or deformation artifacts caused by large head motion. The warped images are produced by warping the source image with the motion flow; (\textbf{e}) presents randomly sampled memory channels of our learned memory bank conditioned with implicit-keypoint representations. Examples of generated results with our memory compensation network are shown in (\textbf{f}).}
    \label{fig:mov}
\end{center}
}]

\maketitle

\ificcvfinal\thispagestyle{empty}\fi

\begin{abstract}
Talking head video generation aims to animate a human face in a still image with dynamic poses and expressions using motion information derived from a target-driving video, while maintaining the person's identity in the source image. However, dramatic and complex motions in the driving video cause ambiguous generation, because the still source image cannot provide sufficient appearance information for occluded regions or delicate expression variations, which produces severe artifacts and significantly degrades the generation quality.
To tackle this problem, we propose to learn a global facial representation space, and design a novel implicit identity representation conditioned memory compensation network, coined as MCNet, for high-fidelity talking head generation.~Specifically, we devise a network module to learn a unified spatial facial meta-memory bank from all training samples, which can provide rich facial structure and appearance priors to compensate warped source facial features for the generation. Furthermore, we propose an effective query mechanism based on implicit identity representations learned from the discrete keypoints of the source image. It can greatly facilitate the retrieval of more correlated information from the memory bank for the compensation. Extensive experiments demonstrate that MCNet can learn representative and complementary facial memory, and can clearly outperform previous state-of-the-art talking head generation methods on VoxCeleb1 and CelebV datasets. Please check our \href{https://github.com/harlanhong/ICCV2023-MCNET}{Project}. 

\end{abstract}

\section{Introduction}
In this paper, we aim at addressing the problem of generating a realistic talking head video given one still source image and one dynamic driving video, which is widely known as talking head video generation. A high-quality talking head generation model needs to imitate vivid facial expressions and complex head movements, and should be applicable for different facial identities presented in the source image and the target video. It has been attracting rapidly increasing attention from the community, and a wide range of realistic applications remarkably benefits from this task, such as digital human broadcast, AI-based human conversation, and virtual anchors in films.

\par Significant progress has been achieved on this task in terms of both quality and robustness in recent years. Existing works mainly focus on learning more accurate motion estimation and representation in 2D or 3D to improve the generation. More specifically, 2D facial keypoints or landmarks are learned to model the motion flow (see Fig.~\ref{fig:mov}c) between the source image and any target image in the driving video~\cite{zhao2021sparse, zakharov2019few,hong2022depth}. Some works also consider utilizing 3D facial prior model (\eg~3DMM\cite{blanz1999morphable}) with decoupled expression codes~\cite{zhao2021sparse, zakharov2019few} or learning dense facial geometries in a self-supervised manner~\cite{hong2022depth} to model complex facial expression movements to produce more fine-grained facial generation. However, no matter how accurately the motion can be estimated and represented, highly dynamic and complex motions in the driving video cause ambiguous generation from the source image (see Fig.~\ref{fig:mov}d), because the still source image cannot provide sufficient appearance information for occluded regions or delicate expression variations, which severely produces artifacts and significantly degrades the generation quality. 

\par Intuitively, we understand that human faces are highly symmetrical and structured, and many regions of the human faces are essentially not discriminative. For instance, only blocking a very small eye region of a face image makes a well-trained facial recognition model largely drop the recognition performance~\cite{qiu2021end2end}, which indicates to a certain extent that the structure and appearance representations of human faces crossing different face identities are generic and transferable.~Therefore, learning global facial priors on spatial structure and appearance from all available training face images, and utilizing the learned facial priors for compensating the dynamic facial synthesis is highly potential for high-fidelity talking head generation, while it has been barely explored in existing works.
\par
\begin{figure*}[!t]
  \centering
\includegraphics[width=1\linewidth]{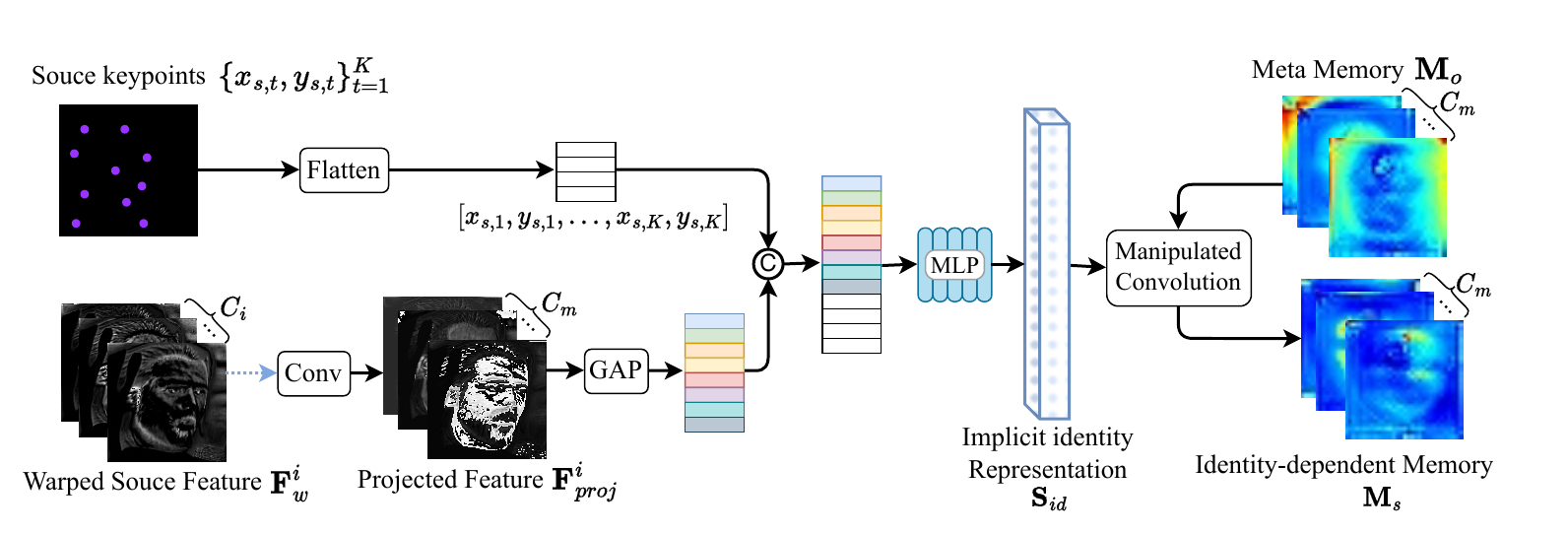}
    \caption{Illustration of the proposed implicit identity representation conditioned memory module (IICM). The symbol \textcircled{c} denotes the concatenation operation, and the ``GAP'' and ``Conv'' represent the global average pooling and the convolution layer, respectively. The detailed generation of the projected feature $\mathbf{F}_{proj}^i$ can refer to Fig.~\ref{fig:mcm}. $C_i$ denotes the channel number of the $i$-th level warped feature $\mathbf{F}_w^i$, while $C_m$ is the channel number of our global facial meta-memory bank.}
    \label{fig:iicm}    
\end{figure*}

 In this paper, to effectively deal with the ambiguities in dramatic appearance changes from the still source image, we propose an implicit identity representation conditioned \textbf{M}emory \textbf{C}ompensation \textbf{Net}work, coined as \textbf{MCNet}, to learn and transfer global facial representations to compensate ambiguous facial details for a high-fidelity generation. Specifically, we design and learn a global and spatial facial meta-memory bank. The optimization gradients from all the training images during training contribute together to the updating of the meta memory, and thus it can capture the most representative facial patterns globally. Since the different source face images contain distinct structures and appearances, to more effectively query the learned global meta memory bank, we propose an implicit identity representation conditioned memory module (IICM) (see Fig.~\ref{fig:iicm}). The implicit {identity} representation is learned from both the discrete keypoint coordinates of the source face image that contains the facial structure information, and the warped source feature map that represents facial appearance distribution. Then, we further use it to condition the query on the global facial meta-memory bank to learn a more correlated memory bank for the source, which can effectively compensate the source facial feature maps for the generation. The compensation is then performed through a proposed memory compensation module (MCM) (see Fig.~\ref{fig:mcm}). 

\par We conduct extensive experiments to evaluate the proposed MCNet on two competitive talking head generation datasets (\ie~VoxCeleb~\cite{nagrani2017voxceleb} and CelebV~\cite{wu2018reenactgan}. Experimental results demonstrate the effectiveness of learning global facial memory to tackle the appearance ambiguities in the talking head generation, and also show clearly improved generation results over state-of-the-art methods from both qualitative and quantitative perspectives.

In summary, our main contribution is three-fold:
\begin{itemize}
    \item We propose to learn a global facial meta-memory bank to transfer representative facial patterns to handle the appearance and structure ambiguities caused by highly dynamic generation from a still source image. To the best of our knowledge, it is the first exploration in the literature to model global facial representations to address the ambiguities in talking head generation.
    \item We propose a novel implicit identity representation conditioned memory compensation network (MCNet) for talking head video generation, in which an implicit identity representation conditioned memory module (IICM) and a facial memory compensation module (MCM) are designed to respectively perform the meta-memory query and feature compensation.  
    \item Qualitative and quantitative experiments extensively show the effectiveness of the learned meta memory bank for addressing the ambiguities in generation, and our framework establishes a clear state-of-the-art performance on the talking head generation. The generalization experiment also shows that the proposed approach can effectively boost the performance of different talking head generation frameworks.
\end{itemize}

\begin{figure*}[ht]
  \centering
    \includegraphics[width=1\linewidth]{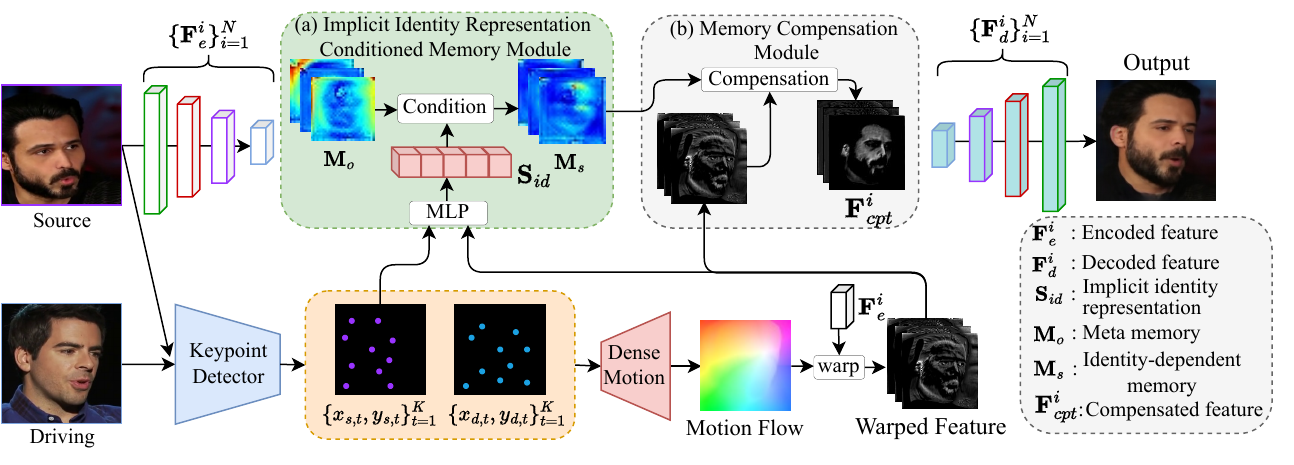}
    \caption{An overview of the proposed MCNet. It contains two designed modules to compensate the source facial feature map: (i) The implicit identity representation conditioned memory module (\textbf{IICM}) learns a global facial meta-memory bank, and 
    an implicit identity representation from facial keypoint coordinates of the source image, which conditions on the query of the learned meta-memory bank, to obtain more structure-correlated facial memory to the warped source feature map for compensation; 
    (ii) The memory compensation module (\textbf{MCM}) designs a dynamic cross-attention mechanism to perform a spatial compensation for the warped source feature map for the generation. 
    }
    \label{fig:framework}    
\end{figure*}
\section{Related Works}
\noindent\textbf{Talking Head Video Generation.} Talking Head video Generation can be mainly divided into two strategies: image-driven and audio-driven generation. For the image-driven strategy, researchers aim to capture the expression of a given driving image and aggregate the captured expression with the facial identity from a given source image. Several approaches~\cite{yao2020mesh,wu2021f3a,wang2021safa} utilized a 3DMM regressor~\cite{tran2018nonlinear,zhu2017face} to extract an expression code and an identity code from a given face, and then respectively combine them from different faces to generate a new face. Also, some other works~\cite{tripathy2021facegan,ha2020marionette,zakharov2020fast,zakharov2019few,zhao2021sparse} utilized facial landmarks detected by a pretrained face model~\cite{guo2019pfld} to act as anchors of the face. Then, the facial motion flow calculated from the landmarks is transferred from a driving face video. However, their motion flow suffers from error accumulation caused by inaccuracy of the pretrained model. To overcome this limitation, the keypoints are learned in an unsupervised fashion~\cite{siarohin2019first,hong2022depth,wang2021one,liu2021self,zhao2022thin} to better represent the motion of the face with carefully designed mechanisms for modeling the motion transformations between two sets of keypoints. Audio-driven talking head generation~\cite{ji2022eamm,lu2021live,wu2021imitating,ji2021audio} is another popular direction on this topic, as audio sequences do not contain information of the face identity, and is relatively easier to disentangle the motion information from the input audio. Liang~\etal~\cite{liang2022expressive}~explicitly divide the driving audio into granular parts through delicate priors to control the lip shape, face pose, and facial expression.

In this work, we focus on the image-driven talking head generation. In contrast to previous image-driven works, we aim at learning global facial structure and appearance priors through a well-designed memory-bank network, which can effectively compensate intermediate facial features and produce higher-quality generation on ambiguous regions caused by large head motion.

\noindent\textbf{Memory Bank Learning.} Introducing an external memory or prior component is popular because of its flexible capability of storing, abstracting, and organizing long-term knowledge into a representative form. Recently, the memory bank has shown its powerful capabilities in learning and reasoning for addressing several challenging tasks, \eg~image processing~\cite{yoo2019coloring,huang2021memory}, video object detection~\cite{sun2021mamba}, and image caption~\cite{fei2021memory}. As an earlier work, \cite{weston2014memory} proposes a memory network, which integrates inference components within a memory bank that can be read and written to memorize supporting facts from the past for question answering. 
Xu~\etal~\cite{xu2021texture} use the texture memory of patch samples extracted from unmasked regions to inpaint missing facial parts. \cite{wu2022showface} proposes a memory-disentangled refinement network for coordinated face inpainting in a coarse-to-fine manner.

\par In contrast to these previous works, to the best of our knowledge, we are the first to propose learning a global facial meta-memory bank to deal with ambiguous generation issues in the task of talking head generation. We also accordingly design a novel implicit identity representation conditioned memory query mechanism and a memory compensation network to effectively tackle the issues. 

\section{The Proposed Approach}
\subsection{Overview}
\label{sec:overview}
An overview of our proposed implicit identity representation conditioned memory compensation network for talking head generation is depicted in Fig.~\ref{fig:framework}. It can be divided into three parts: (i) The keypoint detector and the dense motion network. Initially, the keypoint detector receives a source image $\mathbf{S}$ and a driving frame $\mathbf{D}$ to predict $K$ pairs of keypoints, \ie $\{x_{s,t}, y_{s,t}\}_{t=1}^K$ and $\{x_{d,t}, y_{d,t}\}_{t=1}^K$ on the source and target, respectively. With the keypoints generated from the driving frame and the source image, the dense motion network estimates the motion flow $A_{\mathbf{S} \leftarrow \mathbf{D}}$ between these two; (ii) The designed implicit identity representation conditioned memory module (IICM). We first leverage the estimated motion flow $A_{\mathbf{S} \leftarrow \mathbf{D}}$ to warp the encoded feature $\mathbf{F}_e^i$ in the $i$-th layer, resulting in a warped feature $\mathbf{F}_w^i$. The warped feature $\mathbf{F}_w^i$ and the source keypoints are then fed into the IICM module to encode an implicit identity representation, which will condition on the query of the meta memory $\mathbf{M}_o$ to produce a source-identity-dependent memory bank $\mathbf{M}_s$; (iii) The memory compensation module (MCM). After obtaining $\mathbf{M}_s$, we utilize a dynamic cross-attention mechanism to compensate the warped source feature map spatially in the MCM module, and then output a compensated feature map  $\mathbf{F}_{cpt}^i$. Finally, our decoder utilizes all the $N$ feature maps \ie $\{\mathbf{F}_{cpt}^i\}_{i=1}^N$, to produce the final image $\mathbf{I}_{rst}$. 
In the following, we will show how to learn our memory bank in the IICM and how it is utilized in the MCM for generation-feature compensation.

\subsection{Learning Implicit Identity Representation Conditioned Global Facial Meta-memory}
\label{sec:iicm}
We first aim at learning a global meta-memory bank to model facial structure and appearance representations from the whole face dataset. As different human faces have distinct structures and appearances, and thus using the whole learned facial meta-memory bank to directly compensate different source faces is inflexible. To handle this issue, we further learn an implicit identity representation from discrete keypoint coordinates of the source face and the corresponding warped source feature map. It is then used to condition on the query of 
the global meta-memory bank and obtain source-identity-dependent feature memory, which compensates the warped source feature map for generation.

\noindent\textbf{Global facial meta-memory.} In this work, we first aim to learn a global facial meta-memory bank to store the global and representative facial appearance and structure priors from all the training data available. We initialize a meta-memory bank $\mathbf{M}_o$ as a cube tensor with a shape of $C_m \times H_m\times W_m$ instead of a vector~\cite{esser2021taming}. Moreover, the multiple channels hold sufficient capacity for the meta-memory bank to learn different facial structures and appearances (see Fig.~\ref{fig:meta-bank}). As many regions of the human faces are not discriminative and transferable, we can utilize the global facial priors learned in the meta-memory to compensate ambiguous regions in the generated faces. The meta memory bank is automatically updated by the optimization gradients from all the training images during the training stage, based on an objective function described in Eq.~\ref{eq:con}. In this way, the facial prior learned in the meta memory is global rather than conditioned on any specific input sample, providing highly beneficial global patterns for  compensating face generation.

\begin{figure*}[ht]
  \centering
\includegraphics[width=1\linewidth]{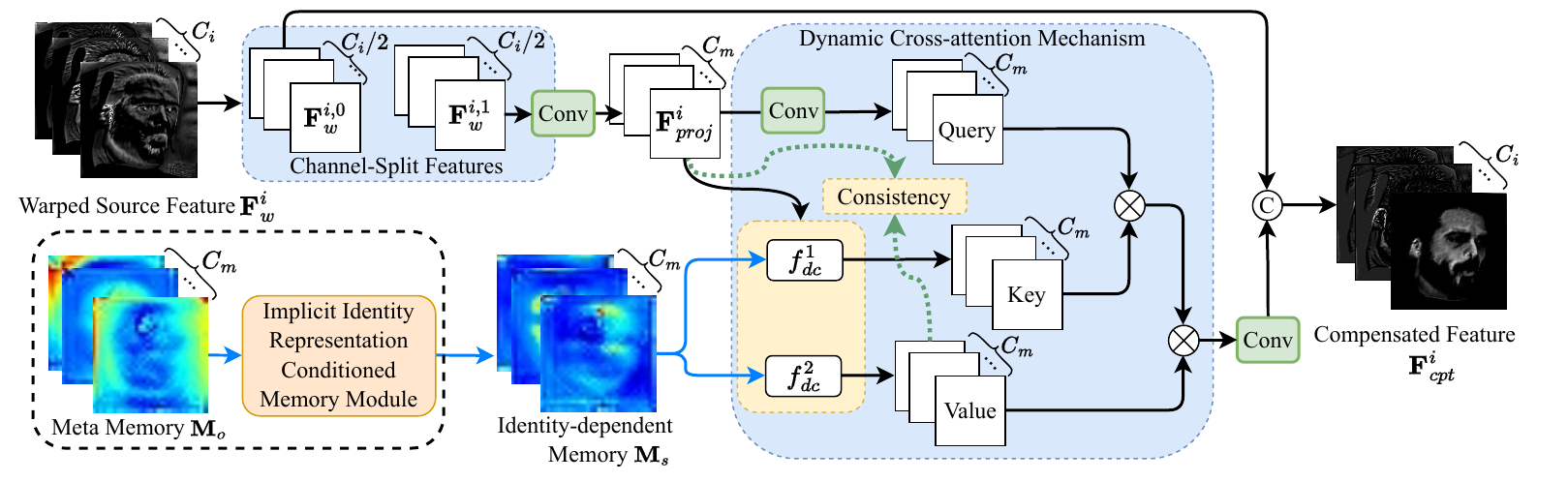}
    \caption{The illustration of the memory compensation module (MCM). The symbol $\bigotimes$ denotes matrix multiplication, and $f_{dc}^1$ and $f_{dc}^2$ are dynamic convolution layers~\cite{chen2020dynamic}, whose kernel weights are estimated by the projected feature $\mathbf{F}_{proj}^i$. The \textcircled{c} represents the concatenation operation, and the ``Conv'' denote a convolution layer. $C_i$ is the channel number of the $i$-th level feature in our autoencoder framework, while $C_m$ is the channel number of the memory bank.}
    \label{fig:mcm}  
\end{figure*}

\noindent\textbf{Implicit identity representation learning.} In our framework, the detected facial keypoints are used to learn motion flow for feature warping. The facial keypoints implicitly contain the structure information of the face because of their structural positions~\cite{siarohin2019first, tao2022structure}.
Therefore, we utilize both the source keypoint coordinates $\{x_{s,t}, y_{s,t}\}_{t=1}^K$ and its corresponding warped feature $\mathbf{F}_w^i$ that provides additional appearance constraints to learn an implicit {identity} representation of the source face. The reason for learning on the source is that we need to compensate the warped facial feature map with the identity of the source.
As shown in Fig.~\ref{fig:iicm}, we first utilize a global average pooling function $\mathcal{P}_{F}$ to squeeze the global spatial information of the projected feature $\mathbf{F}_{proj}^i$ that is produced from the warped feature $\mathbf{F}_{w}^i$ (see Fig.~\ref{fig:mcm}), into a channel descriptor. It is then concatenated with 
flattened and normalized keypoint coordinates, and fed into an MLP mapping network $\mathcal{F}_{mlp}$ to learn an implicit identity representation of the source image. 
We have:
\begin{align*}
    \mathbf{S}_{id} = 
\mathcal{F}_{mlp}\left(\left[\mathcal{P}_{F}(\mathbf{F}_{proj}^i),  [x_{s,1},y_{s,1},\dots,x_{s,K},y_{s,K}]\right]\right), 
\end{align*}
where the operator $[\cdot,\cdot]$ indicates the concatenation operation, and $\mathbf{S}_{id}$ is the learned implicit identity representation of the source face image. 

\noindent\textbf{implicit identity representation conditioned meta-memory learning.}
As discussed before, human faces present distinct structures and appearances. To generate a more correlated facial memory for compensating the source feature map, we utilize the learned implicit identity representation $\mathbf{S}_{id}$ to condition on the retrieval of our global facial meta-memory $\mathbf{M}_o$, which produces an identity-dependent facial memory $\mathbf{M}_s$ for each source face image.
Inspired by the style injection in StyleGANv2~\cite{karras2020analyzing}, we utilize the implicit identity representation $\mathbf{S}_{id}$ to manipulate a $3\times 3$ convolution layer to produce a conditioned facial memory: $\omega_{ijk}^\prime = s_i*\omega_{ijk}$ and $\omega_{ijk}^{\prime\prime} = \omega_{ijk}^\prime  \sqrt{\sum_{i,k}(\omega_{ijk}^\prime)^2+\epsilon}$,
where $\omega$ is the weight of the convolution kernel; $\epsilon$ is a small constant to avoid numerical issues; $s_i$ is the $i$-th element in the learned implicit identity representation $\mathbf{S}_{id}$, and $j$ and $k$ enumerate the output feature maps and spatial footprint of the convolution, respectively. Finally, we obtain the learned source-dependent facial memory:
\begin{equation}
\label{eq:manipulatedC}
    \mathbf{M}_s = \mathcal{F}_{C_{\mathbf{\omega}^{\prime\prime}}}(\mathbf{M}_o)
\end{equation}
where the $\mathcal{F}_{C_{\mathbf{\omega}^{\prime\prime}}}$ is the manipulated convolution layer parameterized by $\mathbf{\omega}^{\prime\prime}$. With the identity-independent memory $\mathbf{M}_s$, each warped source feature map can be compensated by the source-correlated facial priors to have a more effective face generation. 

\subsection{Global Memory Compensation and Generation}
\label{sec:mcm}
The warped source feature map typically contains ambiguity for the generation, especially when the warping is performed under large head motion or occlusion. Thus, we propose to inpaint those ambiguous features via compensating the warped source facial feature maps. To this end, we design a memory compensation module (MCM) as shown in Fig.~\ref{fig:mcm}, to refine the warped feature $\mathbf{F}_w^i$ via the learned source-identity-dependent facial memory bank $\mathbf{M}_s$.

\noindent\textbf{Projection of warped facial feature.}
To maintain better the identity information in the source image while compensating the warped source feature map, we employ a channel-split strategy to split the warped feature $\mathbf{F}_w^i$ into two parts along the channel dimension, \ie~$\mathbf{F}_w^{i,0}$ and $\mathbf{F}_w^{i,1}$. The first half of channels $\mathbf{F}_w^{i,0}$ are directly passed through for contributing the identity preserving, while the rest half of channels $\mathbf{F}_w^{i,1}$ are modulated by the source identity-dependent memory bank $\mathbf{M}_s$, to refine the ambiguities. After splitting, we employ a $1\times 1$ convolution layer on $\mathbf{F}_w^{i,1}$ to change the channel number, resulting in a projected feature map $\mathbf{F}_{proj}^{i}$. 

\noindent\textbf{Warped facial feature compensation.} We adopt a dynamic cross-attention mechanism to compensate the warped source feature map spatially. Specifically, we employ the identity-dependent memory $\mathbf{M}_s$ to produce the Key $\mathbf{F}_K^i$ and Value $\mathbf{F}_V^i$ via two dynamic convolution layers (\ie~$f_{dc}^1$, $f_{dc}^2$) conditioned on the projected feature $\mathbf{F}_{proj}^{i}$. In this way, the generated Key and Value are identity-dependent and capable of providing useful context information. In the meanwhile, we perform a non-linear projection to map $\mathbf{F}_{proj}^{i}$ into a query feature $\mathbf{F}_Q^i$ by a $1\times 1$ convolution layer followed by a ReLU layer. 
Then, we perform cross attention to reconstruct a more robust feature $\mathbf{F}_{ca}^i$ as:
\begin{equation}
    \mathbf{F}_{ca}^i=\mathcal{F}_{C_{1\times 1}}\left({\rm Softmax}\left({\mathbf{F}_Q^i}^T \times \mathbf{F}_K^i\right) \times \mathbf{F}_V^i\right),
\end{equation}
where ``{\rm Softmax}'' denotes the softmax operator, while the $\mathcal{F}_{C_{1\times 1}}$ is a $1\times 1$ convolution layer to change the channel number of the cross-attention output. ``$\times$'' denotes a matrix multiplication. As shown in Fig.~\ref{fig:mcm}, to maintain the identity of the source image, we concatenate the cross-attention features $\mathbf{F}_{ca}^i$ with the first-half channels $\mathbf{F}_w^{i,0}$:
\begin{equation}
\label{eq:f3c}
 \mathbf{F}_{cpt}^i = {\rm Concat}[\mathbf{F}_{ca}^i, \mathbf{F}_w^{i,0}],
\end{equation}
where the ${\rm Concat}[\cdot,\cdot]$ represents a concatenation operation. 
As a result, the final output feature map $\mathbf{F}_{cpt}^i$ can effectively benefit and incorporate the learned facial prior information~\cite{wang2021towards} from the memory, modulated by the dynamic cross-attention mechanism.

\noindent\textbf{Regularization on consistency.} 
To learn the global facial appearance and structure representations from the input training face images, we need to make the learning of the meta-memory constrained by every single image in the training data. Simply but effectively, we enforce the consistency between the projected feature $\mathbf{F}_{proj}^i$ from the current training face image, and the value feature $\mathbf{F}_V^i$ from the global meta memory: 
\begin{equation}
\mathcal{L}_{con} = ||\mathbf{F}_V^i - de(\mathbf{F}_{proj}^i)||_1,
\label{eq:con}
\end{equation}
where the $de(\cdot)$ indicates a gradient detach function and $||\cdot||_1$ is $\mathcal{L}_1$ loss. By using this function, the regularization enforces the consistency into the learning of the global meta-memory, while not affecting the learning of the source image features. This can guarantee the stability of training the overall generation framework. 
The above equation also makes sure that the optimization gradients from all the face images during the training state contribute together to the updating of the memory bank, and thus it can capture global facial representations for the generation compensation.

\par\noindent\textbf{Multi-layer generation.} 
Following previous works~\cite{zhao2022thin} considering multi-scale features in generation,
we also perform memory compensation for feature maps of multiple layers to enhance facial details. As shown in Fig.~\ref{fig:framework}, we utilize the motion flow $A_{\mathbf{S}\leftarrow \mathbf{D}}$ to warp the encoded feature $\{\mathbf{F}_e^i\}_{i=1}^N$ in each layer to produce warped features $\{\mathbf{F}_w^i\}_{i=1}^N$. For each warped feature $\mathbf{F}_w^i$, we feed it into our designed IICM and MCM modules sequentially to produce the compensated feature maps $\{\mathbf{F}_{cpt}^i\}_{i=1}^N$. In the decoding process, we treat the $\mathbf{F}_{cpt}^1$ as $\mathbf{F}_{d}^1$ and then the $\mathbf{F}_{d}^2$ is generated by $\mathbf{F}_{d}^1$ through an upsampling layer. At the $i$-th level ($i>1$), the output compensated feature map $\mathbf{F}_{cpt}^i$ is concatenated with the decoded feature $\mathbf{F}_d^i$ at the same level to produce a decoded feature $\mathbf{F}_d^{i+1}$. 
Finally, we input the concatenation of $\mathbf{F}_d^N$ and $\mathbf{F}_{cpt}^N$ into a convolution layer followed by a Sigmoid unit to generate the final facial image $\mathbf{I}_{rst}$. Each layer shares the same meta memory $\mathbf{M}_o$.

\begin{figure*}[ht]
  \centering
    \includegraphics[width=0.98\linewidth]{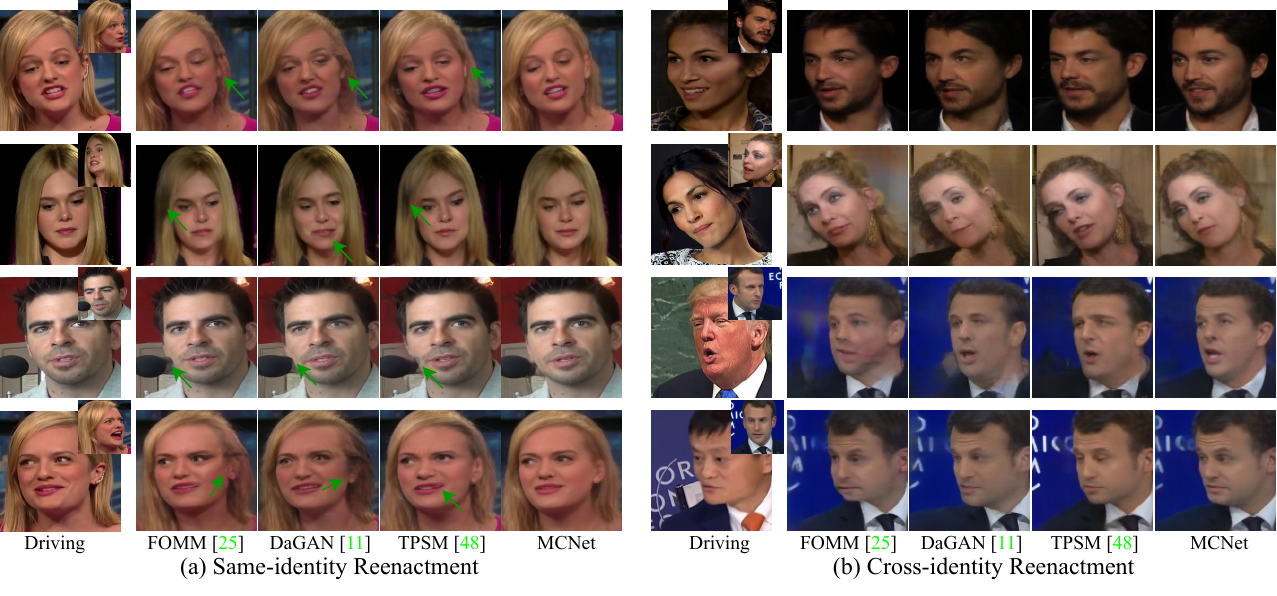}
    \caption{Qualitative comparisons of (a) same-identity reenactment and (b) cross-identity reenactment on the VoxCeleb1 (the first two rows) and CelebV dataset (the last two rows). Our method shows higher-fidelity generation compared to the state-of-the-arts. {Zoom in for best view.}}
    \label{fig:comparison}   
\end{figure*}

\begin{table*}[t]
  \centering
  \resizebox{0.95\linewidth}{!}{
       
        \begin{tabular}{l|cccccc||cc|cc}
        \toprule
        \multirow{3}*{Model} & \multicolumn{6}{c||}{(a) Results of Same-identity Reenactment}  & \multicolumn{4}{c} {(b) Results of Cross-identity Reenactment}\\
        \cline{2-11}
        &\multicolumn{6}{c||} {VoxCeleb1}& \multicolumn{2}{c|} {VoxCeleb1} & \multicolumn{2}{c} {CelebV1}  \\
         & SSIM (\%) $\uparrow$ & PSNR $\uparrow$  &  LPIPS $\downarrow$ & $\mathcal{L}_1$ $\downarrow$ & AKD $\downarrow$ &  AED $\downarrow$ & AUCON $\uparrow$ & PRMSE $\downarrow$ & AUCON $\uparrow$ & PRMSE $\downarrow$\\
        \midrule
        X2face \cite{wiles2018x2face} &71.9 & 22.54  & - & 0.0780&7.687&0.405 & - & - &  0.679 & 3.62\\
        marioNETte \cite{ha2020marionette} &75.5 & 23.24 & - & - &- & - & - & - &  0.710 & 3.41\\
      FOMM \cite{siarohin2019first}) & 72.3 & 30.39  & 0.199  & 0.0430 & 1.294 & 0.140 & 0.882 & 2.824 & 0.667 & 3.90\\
      MeshG \cite{yao2020mesh}& 73.9&30.39 & - & - & - & - & - & -  & 0.709 & 3.41\\
      face-vid2vid \cite{wang2021one}& 76.1& 30.69 & 0.212 & 0.0430 & 1.620 & 0.153 & 0.839 & 4.398 &  0.805 & 3.15\\
      MRAA \cite{siarohin2021motion}& 80.0  & 31.39  & 0.195 & 0.0375 & 1.296 &0.125 & 0.882 & 2.751 & 0.840 & 2.46  \\
      DaGAN \cite{hong2022depth}& 80.4 &31.22  & 0.185  & 0.0360 &1.279 &0.117 & 0.888 & 2.822 & 0.873 & 2.33 \\
      TPSN \cite{zhao2022thin}& 81.6  & 31.43  & 0.179 &0.0365 &1.233 &0.119 & 0.894 & 2.756 & 0.882 & 2.23\\
        \midrule
        MCNet (Ours) & \textbf{82.5} & \textbf{31.94}  & \textbf{0.174} & \textbf{0.0331} & \textbf{1.203} & \textbf{0.106} & \textbf{0.895} & \textbf{2.641} & \textbf{0.885} & \textbf{2.10} \\
        \bottomrule
        \end{tabular}
}
\caption{Comparisons with state-of-the-art methods on same-identity reenactment on VoxCeleb1 (see Fig.~\ref{fig:comparison}a) and cross-identity reenactment on VoxCeleb1 and CelebV dataset (see Fig.~\ref{fig:comparison}b).}
\label{tab:comparison}
\end{table*}

\begin{figure*}[t]
  \centering
    \includegraphics[width=1\linewidth]{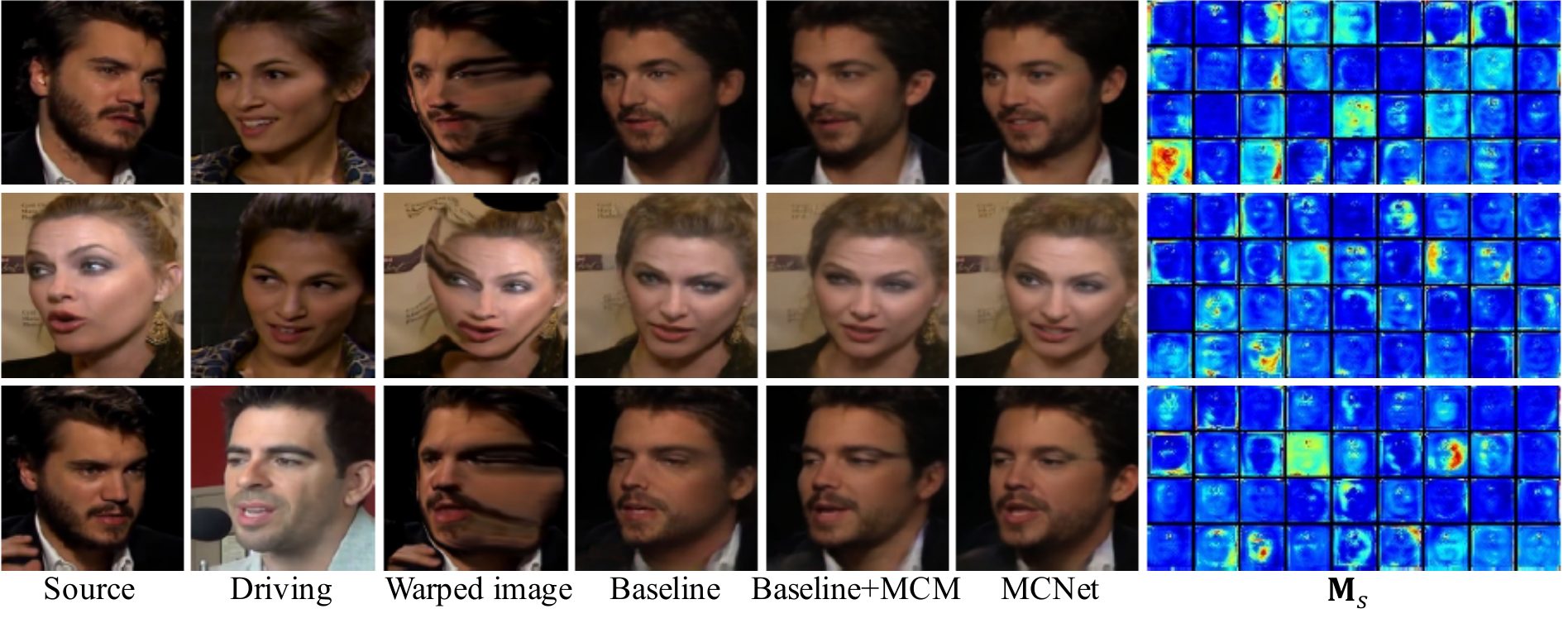}
    \vspace{-18pt}
    \caption{Qualitative ablation studies. The memory compensation module (MCM) and implicit identity representation conditioned memory module (IICM) can both effectively improve the generation performances. The last column verifies that our IICM can learn identity-conditioned memories (\ie~$\mathbf{M}_s$) for the different source face samples.}
    \label{fig:abla}  
\end{figure*}
\begin{table*}[t]
  \centering
  \resizebox{0.85\linewidth}{!}{
        \begin{tabular}{lcccccc}
        \toprule
        Model & SSIM (\%) $\uparrow$ & PSNR $\uparrow$ &  LPIPS $\downarrow$ & $\mathcal{L}_1$ $\downarrow$ & AKD $\downarrow$ &  AED $\downarrow$ \\
        \midrule
        
        Baseline &81.1& 31.70& 0.182&  0.0356& 1.303 & 0.124\\
        Baseline + MCM$^{\text{w/o~Eq.~\ref{eq:f3c}}}$ & 82.0 &31.82 & 0.176& 0.0340& 1.242 & 0.119 \\
        Baseline + MCM & 82.3 & 31.92 &  0.175 &  0.0334& 1.237&0.114 \\
        Baseline + IICM + MCM (MCNet)& \textbf{82.5} & \textbf{31.94} & \textbf{0.174}& \textbf{0.0331} & \textbf{1.203} & \textbf{0.106} \\
      
        \midrule
        \midrule
        FOMM~\cite{siarohin2019first} & 72.3 & 30.39 & 0.199  & 0.0430 & 1.294 & 0.140\\
        FOMM+ IICM + MCM & \textbf{81.8} & \textbf{31.73} & \textbf{0.179}  & \textbf{0.0353} & \textbf{1.269} & \textbf{0.119} \\\midrule
        TPSN~\cite{zhao2022thin} & 81.6  & 31.43  & 0.179 &0.0365 &1.233 &0.119\\
        TPSN+ IICM + MCM & \textbf{82.0} & \textbf{31.55} & \textbf{0.175}  & \textbf{0.0356} & \textbf{1.216} & \textbf{0.115} \\
        \bottomrule
        \end{tabular}
}
\caption{Ablation studies: ``Baseline'' indicates the simplest model without the implicit identity representation conditioned memory module (IICM) and memory compensation module (MCM). ``MCM$^{\text{w/o~Eq.~\ref{eq:f3c}}}$'' indicates that we use the entire warped feature to generate the projected feature $\mathbf{F}_{proj}^i$ without using the channel split. The compensated feature $\mathbf{F}_{cpt}^i$ for generation is thus directly from the output of cross-attention query (\ie~$\mathbf{F}_{ca}^i$) of the meta memory $\mathbf{M}_o$.}
\label{tab:abla}
\end{table*}
\subsection{Training}
\label{sec:training}
We train the proposed MCNet by minimizing several optimization losses. Similar to FOMM~\cite{siarohin2019first}, we leverage the perceptual loss $\mathcal{L}_{P}$ to minimize the gap between the model output and the driving image, and equivariance loss $\mathcal{L}_{eq}$ to learn more stable keypoints. Additionally, we also adopt the keypoints distance loss $\mathcal{L}_{dist}$~\cite{hong2022depth} to avoid the detected keypoints crowding around a small neighborhood. The $\mathcal{L}_{con}$ is the regularization consistency loss described in Eq.~\ref{eq:con}. The overall loss function is written as follows:
\begin{equation}
    \mathcal{L} =\lambda_{P}\mathcal{L}_{P}+\lambda_{eq}\mathcal{L}_{eq} +\lambda_{dist}\mathcal{L}_{dist}+ \lambda_{con}\mathcal{L}_{con},
\end{equation}
where the $\lambda_{P}$, $\lambda_{eq}$, $\lambda_{dist}$ and $\lambda_{con}$ are the hyper-parameters to allow for balanced learning from these losses. More details about these losses are described in Supplementary.
\section{Experiments}
In this section, we present quantitative and qualitative experiments to validate the effectiveness of our MCNet. 
\subsection{Datasets and Metrics}
\noindent\textbf{Dataset}. We evaluate our MCNet on two talking head generation datasets, \ie~VoxCeleb1~\cite{nagrani2017voxceleb} and CelebV~\cite{wu2018reenactgan} dataset. We follow the sampling strategy for the test set in DaGAN~\cite{hong2022depth} for evaluation. Following DaGAN, to verify the generalization ability, we apply the model trained on VoxCeleb1 to test on CelebV.

\noindent\textbf{Metrics.} We adopt the structured similarity (\textbf{SSIM}), peak signal-to-noise ratio (\textbf{PSNR}), and $\mathbf{\mathcal{L}}_1$ distance to measure the low-level similarity between the generated image and the driving image. Following the previous works~\cite{siarohin2019first}, we utilize the Average Euclidean Distance (\textbf{AED}) to measure the identity preservation, and Average Keypoint Distance (\textbf{AKD}) to evaluate whether the motion of the input driving image is preserved. We also adopt the \textbf{AUCON} and \textbf{PRMSE}, similar to~\cite{hong2022depth}, to evaluate the expression and head poses in cross-identity reenactment.
\subsection{Comparison with state-of-the-art methods}
\noindent\textbf{Same-identity reenactment.} In Table~\ref{tab:comparison}(a), we first compare the synthesised results for the setup in which the source and the driving images share the same identity. It can be observed that our MCNet obtains the best results compared with other competitive methods. Specifically, compared with FOMM~\cite{siarohin2019first} and DaGAN~\cite{hong2022depth}, which adopt the same motion estimation method as ours, our method can produce higher-quality images ($72.3\%$ of FOMM vs $82.5\%$ of ours, resulting in a $10.2\%$ improvement on the SSIM metric), which verifies that introducing the global memory mechanism can indeed benefit the image quality in the generation process. Regarding motion animation and identity preservation, our MCNet also achieves the best results (\ie $1.203$ on AKD and 0.$106$ on AED), showing superior performance on the talking head animation. 
Moreover, we show several samples in Fig.~\ref{fig:comparison}(a), and the face samples in Fig.~\ref{fig:comparison}(a) contain large motions (the first, the third, and the last row) and object occlusion (the second row). From Fig.~\ref{fig:comparison}(a), our model can effectively handle these complex cases and produces more completed image generations compared with the state-of-the-art competitors.

\noindent\textbf{Cross-identity reenactment.} We also perform experiments on the VoxCeleb1 and CelebV datasets to conduct the task of the cross-identity face motion animation, in which the source and driving images are from different people. The results compared with other methods are reported in Table~\ref{tab:comparison}. 
Our MCNet outperforms all the other comparison methods. Regarding the head pose imitation, our MCNet can produce the face with a more accurate head pose (\ie 2.641 and 2.10 for VoxCeleb1 and CelebV, respectively, on the PRMSE metric). 
We also present several samples of results with the VoxCeleb1 dataset in Fig.~\ref{fig:comparison}(b). It is clear to observe that our MCNet can mimic the facial expression better than the other methods, such as the smiling countenance shown in the first row. For the unseen person in the CelebV dataset, \eg~the last two rows in Fig.~\ref{fig:comparison}(b), our method can still produce a more natural generation, while the results of other methods contain more obvious artifacts. All of these results verify that the feature compensated by our 
learned memory can produce better results.
\subsection{Ablation Study}
\begin{figure}[t]
  \centering    \includegraphics[width=1\linewidth]{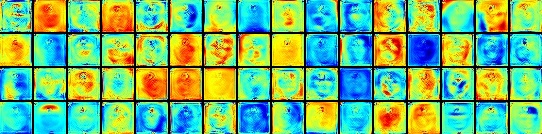}
    \caption{The visualization of randomly selected channels of the global meta memory $\mathbf{M}_o$. It can be observed that our meta-memory learns very diverse facial representations.}
    \vspace{-10pt}
    \label{fig:meta-bank}  
\end{figure}

In this section, we perform ablation studies to demonstrate the effectiveness of the proposed implicit identity representation conditioned memory module (IICM) and memory compensation module (MCM). We report their quantitative results in Table~\ref{tab:abla} and the qualitative results in Fig.~\ref{fig:abla}. Our baseline is the model without IICM and MCM modules. The ``Baseline + MCM'' means that we drop the IICM module and replace the identity-independent  memory $\mathbf{M}_s$ with the meta memory $\mathbf{M}_o$ shown in Fig.~\ref{fig:mcm}. 

\noindent\textbf{Effect of meta memory learning.}
We first visualize the learned meta memory in Fig.~\ref{fig:meta-bank}, which aims to learn the globally representative facial appearance and structure patterns. In Fig.~\ref{fig:meta-bank}, we visualize partial channels of the facial meta-memory bank. It can be observed that these channels represent faces with distinct appearances, structures, scales, and poses, which are very informative and clearly beneficial for facial compensation and generation, confirming our motivation of learning global facial representations to tackle ambiguities in the talking head generation.

\noindent\textbf{Effect of memory compensation.}
In Table~\ref{tab:abla} and Fig.~\ref{fig:abla}, the proposed memory compensation module can effectively improve the generation quality of human faces. From Tab.~\ref{tab:abla}, we observe that adding the memory compensation module (MCM) can consistently boost the performance via a comparison between ``Baseline+MCM'' and ``Baseline'' (82.3\% vs.~81.1\% on SSIM).
In Fig.~\ref{fig:abla}, we can also see that the variant ``Baseline+MCM'' compensates the warped image better than the ``Baseline'', \eg~the face shape in the second row and the mouth shape in the third row. Additionally, we also conduct an ablation study to verify the feature channel split strategy discussed in Sec.~\ref{sec:mcm}. The results of ``Baseline + ${\rm MCM^{w/o\,Eq.\ref{eq:f3c}}}$'' indicate that the channel split can slightly improve the performance. All these results demonstrate that learning a global facial memory can indeed effectively compensate the warped facial feature map to produce higher-fidelity results for the talking head generation.

\noindent\textbf{Effect of implicit identity representation conditioned memory learning.}
To verify the effectiveness of the proposed implicit identity representation conditioned memory module (\ie~IICM introduced in Sec.~\ref{sec:iicm}), we show the randomly sampled channels of the conditioned memory bank in Fig~\ref{fig:abla}. As shown in the last column in Fig~\ref{fig:abla}, the IICM produces an identity-dependent memory bank for the input source images. By deploying the IICM module, our MCNet can generate highly realistic-looking images compared with ``Baseline+MCM'', verifying that the learned memory conditioned on the input source provides a more effective compensation on the warped source feature map for the talking face generation. 

\begin{figure}[h]
  \centering
    \includegraphics[width=1\linewidth]{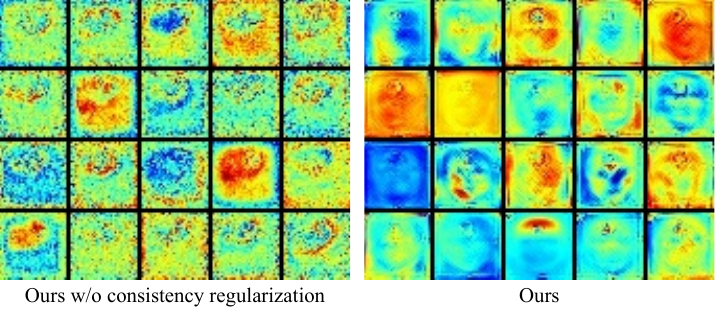}
    \vspace{-16pt}
    \caption{Visualization of selected channels of the meta memory from our full method and the ablation method (\emph{i.e.}, w/o consistency regularization introduced in Eq.~\ref{eq:con}).}
    \label{fig:mb-compare-wo-reg}   
\end{figure}

{
\noindent\textbf{The effect of consistency regularization in Eq.~\ref{eq:con}.} Eq.~\ref{eq:con} we introduced ensures that the optimization gradients across all facial images collaborate in updating the memory bank, capturing global facial representations to improve face generation. Without it, as shown in Fig.~\ref{fig:mb-compare-wo-reg}, the facial memory bank clearly learns less discriminative face patterns and more noisy representations.}

\begin{figure}[t]
  \centering
    \includegraphics[width=1\linewidth]{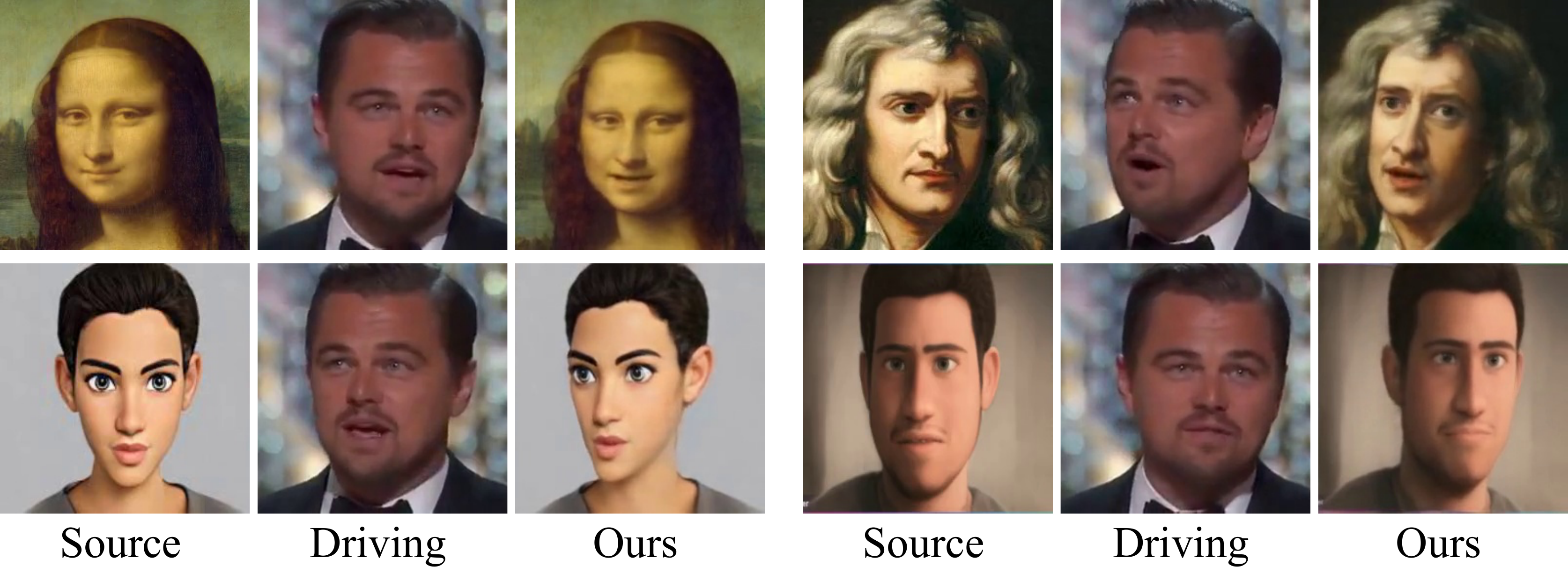}
    \caption{Qualitative results of out-of-domain generation.}
    \label{fig:out-of-domain}     
\end{figure}

\noindent\textbf{Generalization experiment.} Importantly, we also embed the proposed MCM and IICM modules into different representative talking head generation frameworks, including FOMM~\cite{siarohin2019first} and TPSM~\cite{zhao2022thin}, to verify our designed memory mechanism can be flexibly generalized to existing models. As shown in Table~\ref{tab:abla}, the TPSM, which has a different motion estimation method compared to ours, with a deployment of our proposed memory modules, can achieve a stable improvement. The ``FOMM+IICM+MCM'' can also gain a significant improvement on SSIM compared with the pioneering work ``FOMM''. These results demonstrate the transferability and generalization capabilities of the proposed method. {Additionally, we show our generation results on some out-of-domain samples in Fig.~\ref{fig:out-of-domain}, to verify the out-of-domain generation capabilities of our model. As shown in Fig.~\ref{fig:out-of-domain}, our method is able to effectively modify the expression of the oil-painted and cartoon faces.}

\section{Conclusion}
In this paper, we present an implicit identity representation conditioned memory compensation network (MCNet) to globally learn representative facial patterns to address the ambiguity problem caused by the dynamic motion in the talking head video generation task. MCNet utilizes a designed implicit identity representation conditioned memory module to learn the identity-dependent facial memory, which is further used to compensate the warped source feature map by a proposed memory compensation module. Extensive results clearly show the effectiveness of learning global facial meta-memory for the task, producing higher-fidelity results compared with the state-of-the-arts. 

\section{Acknowledgement}
This research is supported in part by HKUST-SAIL joint research funding, the Early Career Scheme of the Research Grants Council (RGC) of the Hong Kong SAR under grant No. 26202321, and HKUST Startup Fund No. R9253.

\appendix
\section*{\LARGE{Appendix}}\label{s:appendix}
\input{appendix}

\newpage
{\small
\bibliographystyle{ieee_fullname}
\bibliography{egbib}
}

\end{document}

%% file: appendix.tex
\section{More implementation details}
The keypoint and dense motion estimation follow FOMM~\cite{siarohin2019first}. Specifically, we extract each frame from the driving video as a driving image and input it into the MCNet model with the source image. 
The source image and driving video share the same identity in the training stage, so the sampled driving frame can be used as the ground-truth of a generated source-identity image.
To optimize the training objectives, we set $\lambda_{rec} = 10$, $\lambda_{eq} = 10$, $\lambda_{dist} = 10$, and $\lambda_{con} = 10$. The number of keypoints is set to 15, which is the same as that of DaGAN~\cite{hong2022depth}. 
In the training stage, we employ 8 RTX 3090 GPUs to run the model for 100 epochs in an end-to-end manner, and it costs about 12 hours in total. The number of layers (\ie~$N$) of the encoder and the decoder is both set as 4, and the number of keypoints (\ie~$K$) is set as $15$ following~\cite{hong2022depth}. We set the size of the proposed meta memory as $512\times 32\times 32$, \ie~$C_m=512$, $H_m=32$ and $W_m=32$. In the motion-based warping process, for any $\mathbf{F}_e^i$ that has a different spatial size to the motion flow, we employ bilinear interpolation to adjust the spatial size of the motion flow.

\section{Network architecture details of MCNet}
\begin{figure}[h]
  \centering
    \includegraphics[width=1\linewidth]{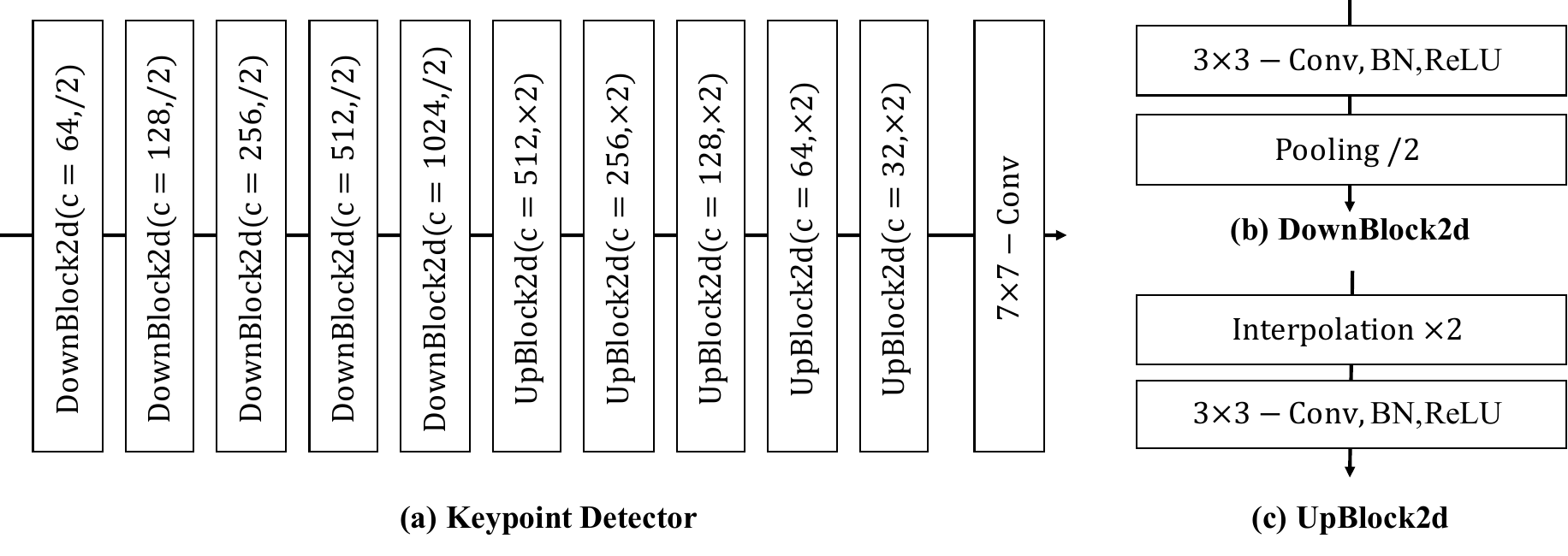}
    \caption{Detailed structure of the keypoint detector. $c$ in each layer indicates the number of output channels.}
    \label{fig:keypoint}       
\end{figure}

The keypoint detector receives an image as input and outputs the $K$ keypoints $\{x_i, y_i\}_{i=1}^K$. The structure of the keypoint detector is illustrated in Fig.~\ref{fig:keypoint}. Here, we adopt the Taylor approximation as FOMM~\cite{siarohin2019first} and DaGAN~\cite{hong2022depth} to compute the motion flow. Thus, the motion estimation is not our focus and we mainly focus on designing our meta memory and its usage for our talking head generation framework.

\section{More details on optimization losses}
\noindent\textbf{Perceptual Loss $\mathcal{L}_p$.} Perceptual loss is a popular objective function in image generation tasks. As introduced in DaGAN~\cite{hong2022depth}, a generated image and its ground-truth, \ie the driving image in the training stage, is downsampled to 4 different resolutions (\ie $256\times 256$, $128\times 128$, $64\times 64$, $32\times 32$), respectively. Then we utilize a pre-trained VGG network~\cite{simonyan2014very} to extract the features from images at each resolution. To simplify, we denote $R_1$, $R_2$, $R_3$, $R_4$ as the features of generated images in different resolutions, respectively, and $G_1$, $G_2$, $G_3$, $G_4$ for the 4 different resolutions of the ground-truth. Then, we measure the $\mathcal{L}_1$ distance between the ground-truth and the generated image by the Perceptual loss defined as follows:
\begin{equation}
    \mathcal{L}_p = \sum_{i=1}^4 \mathcal{L}_1(G_i, R_i)
\end{equation}

 \noindent\textbf{Equivariance Loss $\mathcal{L}_{eq}$.} We employ this loss to maintain the consistency of the estimated keypoints in the images after different augmentations. Per FOMM~\cite{siarohin2019first}, given an image $I$ and its detected keypoints $\{X_i\}_{i=1}^K$ ($X_i \in \mathbb{R}^{1\times 2}$), we first perform a known spatial transformation $T$ on images $I$ and keypoints $\{X_i\}_{i=1}^K$, resulting in a transformed image $I_T$ and transformed keypoints $\{X^T_i\}_{i=1}^K$. Then, we detect keypoints on the transformed image $I_T$, which are denoted as $(\{X_{{I_T},i}\}_{i=1}^K)$. We employ the equivariance Loss on the source image and driving image:
 \begin{equation}
     \mathcal{L}_{eq} = \sum_{i=1}^{K}||X_i^T - X_{{I_T},i}||_1
 \end{equation}

\noindent\textbf{Keypoint distance loss $\mathcal{L}_{dist}$.} We employ the keypoint distance loss as in~\cite{hong2022depth} to penalize the model if the distance between any two keypoints is smaller than a user-defined threshold. Thus, the keypoint distance loss can make the keypoints much less crowded around a small neighbourhood. In one image, for every two keypoints $X_i$ and $X_j$, we then have:
\begin{equation}
    \mathcal{L}_{dist} = \sum_{i=1}^K\sum_{j=1}^K (1-\mathbf{sign}(||X_i-X_j||_1-\alpha)), i\neq j,
\end{equation}
where the $\mathbf{sign}(\cdot)$ represents a sign function and the $\alpha$ is the threshold of the distance, which is $0.2$ in our work.

\section{More details on experiments}
\subsection{Evaluation Metrics}
We mainly consider four important metrics that are widely used in the talking head generation field, \textit{i.e.}, AED, ADK, PRMSE, and AUCON. Specifically, \noindent\textbf{Average euclidean distance (AED)} is an important metric that measures identity preservation in reconstructed video/image. In this work, we use OpenFace~\cite{baltruvsaitis2016openface} to extract identity embeddings from the reconstructed face and the ground truth frame. The MSE loss is used to measure their difference.

\noindent\textbf{Average keypoint distance (ADK).} ADK evaluates the difference between landmarks of the reconstructed faces and the ground truth frames. We extract facial landmarks using a face alignment method~\cite{bulat2017far}. We compute an average distance between the corresponding keypoints. Thus, the AKD mainly measures the ability of the pose imitation.

\noindent\textbf{The root mean square error of the head pose angles (PRMSE).} In this work, we utilize the Py-Feat toolkit\footnote{https://py-feat.org} to detect the Euler angles of the head pose, and then evaluate the pose difference between different identities.

\noindent\textbf{The ratio of identical facial action unit values (AUCON).} 
We first utilize the Py-Feat toolkit to detect the action units of the generated face and the driving face. Then we can calculate the ratio of identical facial action unit values as the AUCON metric.

\subsection{Additional experimental results}
\noindent\textbf{Positional Encoding for keypoints.} The positional encoding method shows its strong power in transformers~\cite{vaswani2017attention, liu2021swin, dosovitskiy2020image} and NeRFs~\cite{mildenhall2020nerf, martin2021nerf,pumarola2021d}. Therefore, we consider applying the positional encoding function\footnote{Here, we use the implementation of https://github.com/yenchenlin/nerf-pytorch} on the keypoints, to produce the implicit identity representation conditioned memory. We show the results in Table.\ref{tab:pe}. From Table~\ref{tab:pe}, we observe that when we apply the position encoding function on keypoints, it cannot bring improvements, and even degrades the model performance if we set the $L$ as 20. Since the keypoints are utilized to estimate the motion flow in the dense motion network, the Euclidean distance between any two keypoints is physically meaningful. Therefore, we suppose that employing the positional encoding on keypoints may affect the motion flow estimation, resulting in an unsatisfactory generation.
\begin{table}[t]
  \centering
  \resizebox{1\columnwidth}{!}{
        \begin{tabular}{ccccccc}
        \toprule
        Model & SSIM (\%) $\uparrow$ & PSNR $\uparrow$ &  LPIPS $\downarrow$ & $\mathcal{L}_1$ $\downarrow$ & AKD $\downarrow$ &  AED $\downarrow$ \\
        \midrule
        Ous w/ pe(10) & 82.4 & 31.91& 0.175&  0.0334 & 1.221 & 0.107\\
        Ous w/ pe(20) & 69.4 & 30.03 & 0.269& 0.0593& 5.544 & 0.268 \\
       
        \midrule
        MCNet& \textbf{82.5} & \textbf{31.94} & \textbf{0.174}& \textbf{0.0331} & \textbf{1.203} & \textbf{0.106} \\
        \bottomrule
        \end{tabular}
}

\caption{The results of applying positional encoding function on keypoints. ``pe(10)'' means that we set the output dimension control factor $\mathbf{L}$ of positional encoding function as 10, and 20 for ``pe(20)''.}
\label{tab:pe}
\end{table}

\begin{table}[t]
  \centering
  \resizebox{1\columnwidth}{!}{
        \begin{tabular}{lcccccc}
        \toprule
        Model & SSIM (\%) $\uparrow$ & PSNR $\uparrow$ &  LPIPS $\downarrow$ & $\mathcal{L}_1$ $\downarrow$ & AKD $\downarrow$ &  AED $\downarrow$ \\
        \midrule

        MCNet (IICM w/o $\mathbf{F}_{proj}^i$) & 82.3 & 31.89& 0.175& 0.0336 & 1.237 & 0.110 \\
        MCNet (IICM w/o keypoints) & 82.4& 31.93& 0.175& 0.0333& 1.227&0.109 \\
      
        MCNet (MCM w/o $f_{dc}^1$, $f_{dc}^2$) & 82.2& 31.89& 0.176& 0.0336& 1.246&0.112 \\
        MCNet (single layer)& 82.3& 31.90& 0.175& 0.0334& 1.235 & 0.108 \\
        \midrule
        MCNet& \textbf{82.5} & \textbf{31.94} & \textbf{0.174}& \textbf{0.0331} & \textbf{1.203} & \textbf{0.106} \\
        \bottomrule
        \end{tabular}
}

\caption{Ablation studies. `IICM w/o $\mathbf{F}_{proj}^i$'' and ``IICM w/o keypoints'' represent that IICM does not use the projected feature $\mathbf{F}_{proj}^i$ or keypoints as input (see Fig.~\ref{fig:iicm}), respectively, to encode implicit identity representation. ``IICM w/o $f_{dc}^1$, $f_{dc}^2$'' indicates that we replace the $f_{dc}^1$ and $f_{dc}^2$ with two normal convolution layers to produce the key and the value in MCM. 
}
\label{tab:iicm-abla}
\end{table}
\noindent\textbf{The input elements in IICM.} We also conduct experiments to investigate the usage of intermediate feature $\mathbf{F}_{proj}^i$ (``IICM w/o $\mathbf{F}_{proj}^i$'') and keypoints (IICM w/o keypoints). The results are shown in Table~\ref{tab:iicm-abla}. The results in the table indicate that these two items are both critical for the generation of the implicit-identity representation conditioned memory bank. We can obtain the best results when we combine them together. 

\noindent\textbf{Single layer vs. multiple layers.} In our work, we deploy the IICM and MCM in each layer to obtain the best results. Also, we investigate the performance of using IICM and MCM in the first layer only. The results ``MCNet (single layer)'' show that the single layer can also obtain similar good results, which can verify the effectiveness of our designed memory mechanism.

\noindent\textbf{The dynamic convolution in MCM.} Besides, we also conduct an ablation study on the dynamic convolution layer in the memory compensation module. We can observe that the dynamic convolution layer can contribute to the final performance, especially for the AKD and AED.
\begin{table}[h]
  \centering
  \resizebox{1\linewidth}{!}{
        \begin{tabular}{lcccccc}
        \toprule
        Model & SSIM (\%) $\uparrow$ & PSNR $\uparrow$ &  LPIPS $\downarrow$ & $\mathcal{L}_1$ $\downarrow$ & AKD $\downarrow$ &  AED $\downarrow$ \\
        \midrule
        FOMM~\cite{siarohin2019first} & 77.19 & 30.71 & 0.257 & 0.0513 & 1.762 &  0.212\\
        MRAA~\cite{siarohin2021motion} & 78.07 & 30.89 & 0.262 & 0.0511 & 1.796 &  0.213 \\
        DaGAN~\cite{hong2022depth} & 79.02 & 30.81 & 0.250 & 0.0483 & 1.865 & 0.341 \\
        TPSM~\cite{zhao2022thin} & 78.22 & 30.63 & 0.254 & 0.0527 & 1.703 & 0.210 \\
        \midrule
        Ours w/o IICM & 78.63 & 31.02 & 0.250& 0.0481 & 1.726&0.199 \\
        Ours & \textbf{79.86} & \textbf{31.18} & \textbf{0.244}& \textbf{0.0470} & \textbf{1.699} & \textbf{0.186} \\
        \bottomrule
        \end{tabular}
}
\caption{State-of-the-art comparison on VoxCeleb2 dataset.}
\label{tab:voxceleb2}
\end{table}

\begin{table}[h]
  \centering
  \resizebox{1\columnwidth}{!}{
        \begin{tabular}{lcccccc}
        \toprule
        Model & SSIM (\%) $\uparrow$ & PSNR $\uparrow$ &  LPIPS $\downarrow$ & $\mathcal{L}_1$ $\downarrow$ & AKD $\downarrow$ &  AED $\downarrow$ \\
        \midrule
        FOMM~\cite{siarohin2019first} & 76.94 & 31.87 & 0.155 & 0.0363 & 1.116 & 0.092 \\
        MRAA~\cite{siarohin2021motion} & 79.36 & 32.32 & 0.156 & 0.0331 & 1.039 &  0.100 \\
        DaGAN~\cite{hong2022depth} & 82.29 & 32.29 & 0.136 & 0.0304 & 1.020 & 0.083 \\
        TPSM~\cite{zhao2022thin} & 86.05 & 32.85 & 0.114 & 0.0264 & 1.015 & 0.072 \\
         \midrule
        Ours w/o IICM & 85.90 & 33.03 & 0.114 & 0.0243& 1.023&0.068\\
        Ours & \textbf{86.45} & \textbf{33.60} & \textbf{0.112}& \textbf{0.0238} & \textbf{0.998} & \textbf{0.064} \\
        \bottomrule
        \end{tabular}
}
\caption{State-of-the-art comparison on HDTF dataset.}

\label{tab:hdtf}
\end{table}

\noindent\textbf{More datasets for evaluation.} To fully verify the superiority of our method, we also compare it with other state-of-the-art methods on two other large-scale datasets, \ie VoxCeleb2~\cite{Chung18b} and HDTF~\cite{zhang2021flow}. We report the results in Table~\ref{tab:voxceleb2} and Table~\ref{tab:hdtf}. From these two tables, we can observe that our method can still obtain the best results compared with the SOTA methods\footnote{These compared methods have officially released code for us to test on these two datasets.}. These results clearly confirm the superiority of our designed method.

\noindent\textbf{Idenetity Preservation.}  In this section, we reorganize the voxceleb1 dataset and divide it into a training set and a test set. These two sets have the same identity space. That is, the identities of test videos also appear in the training videos. We select 500 videos as the test set and the rest as the training set. The experimental results are shown in Table~\ref{tab:id_preservation}. We can observe that our method obtains higher performance under the setting of the testing identities as a part of the training corpus. One possible reason is that our global face meta-memory is learned from the identities in the training set. In this way, it can better compensate for the facial details of those seen identities.

\begin{table}[h]

  \centering
  \resizebox{1\columnwidth}{!}{
        \begin{tabular}{lcccccc}
        \toprule
        Model & SSIM (\%) $\uparrow$ & PSNR $\uparrow$ &  LPIPS $\downarrow$ & $\mathcal{L}_1$ $\downarrow$ & AKD $\downarrow$ &  AED $\downarrow$ \\
        \midrule
        Identities not in the training set & 82.5& 31.94& 0.174&0.0331&1.203&0.106\\
        Identities in the training set & \textbf{83.6}& \textbf{32.38}& \textbf{0.163} &\textbf{0.0319}&\textbf{1.164}&\textbf{0.102}\\
        \bottomrule
        \end{tabular}
}
\caption{Comparison of different variants on HDTF dataset.}
\label{tab:id_preservation}
\end{table}

\noindent\textbf{Video generation demo.} We also provide several video generation demos to show a more detailed comparison qualitatively with the most competitive methods in the literature. From the demo videos, we can observe that our proposed memory compensation network can compensate the regions that do not appear in the source image, with significantly better results than other methods (\eg the ear region in demo2). These demos are attached in Supplementary Material.

\begin{figure}[h]
  \centering
    \includegraphics[width=1\linewidth]{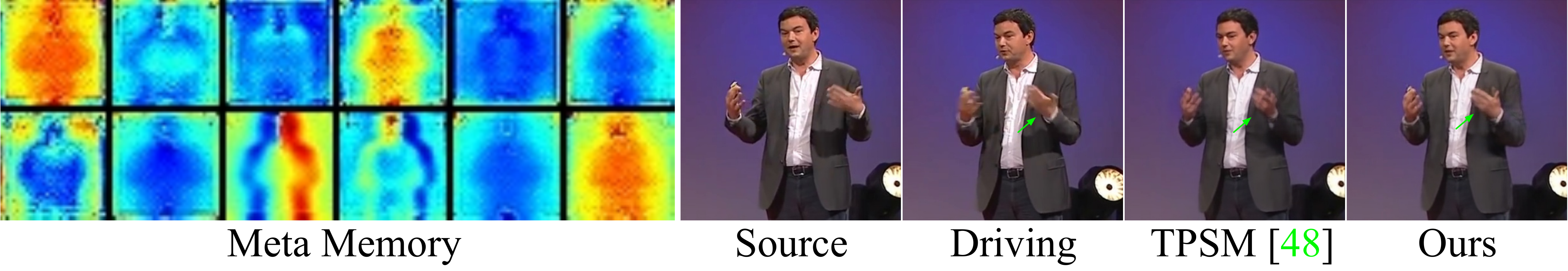}
    \caption{Comparison with TPSM on full-body TED-talks dataset.}
    \label{fig:ted-compare}       
\end{figure}

\noindent\textbf{Comparison on tasks in other domains.} To better verify the generalization ability of our method, we also train our method on TED-talks dataset~\cite{siarohin2021motion}, because the human body is also symmetrical and highly structured. We report the results in Table~\ref{tab:ted}. From the Table~\ref{tab:ted}, our method still obtain the best results among all the compared methods. This generalization experiment verifies that our meta-memory can learn the symmetrical and structured face information to inpaint the generated image. As shown in Fig.~\ref{fig:ted-compare}, our method yields high-quality body details and learns a memory bank with rich and representative body information, which is very useful for the full-body generation. 

\begin{table}[h]

  \centering
  \resizebox{0.86\columnwidth}{!}{
        \begin{tabular}{lccc}
        \toprule
        Model & $\mathcal{L}_1$ $\downarrow$ & (AKD $\downarrow$, MKR $\downarrow$) &  AED $\downarrow$ \\
        \midrule
        FOMM~\cite{siarohin2019first} &0.033&(7.07, 0.014)&0.163 \\
        MRAA~\cite{siarohin2021motion} &0.026&(4.01, 0.012)&0.116 \\
        TPSM~\cite{zhao2022thin} &0.027&(3.39, 0.007)&0.124 \\
        Ours &\textbf{0.023}&(\textbf{2.52}, \textbf{0.006}) & \textbf{0.101}\\
        \bottomrule
        \end{tabular}
}
\caption{State-of-the-art comparison on TED-talks dataset.}
\label{tab:ted}
\end{table}
\begin{table}[h]
  \centering
  \resizebox{1\linewidth}{!}{
        \begin{tabular}{lcccccc}
        \toprule
        Model & SSIM (\%) $\uparrow$ & PSNR $\uparrow$ &  LPIPS $\downarrow$ & $\mathcal{L}_1$ $\downarrow$ & AKD $\downarrow$ &  AED $\downarrow$ \\
        size = 16 & 82.3 & 31.83 & 0.176 & 0.0338& 1.230 & 0.112 \\
        size = 64 & 82.1 & 31.75 & 0.176 & 0.0343& 1.244 & 0.111 \\
        Ours (size = 32) & \textbf{82.5} & \textbf{31.94} & \textbf{0.174}& \textbf{0.0331} & \textbf{1.203} & \textbf{0.106} \\
        \bottomrule
        \end{tabular}
}
\caption{Ablation study on different memory bank sizes.}

\label{tab:size-abla}
\end{table}
\noindent\textbf{The different sizes of the memory bank.} 
We show the ablation studies in Tab.~\ref{tab:size-abla} to inverstigate the effectiveness of different sizes of the memory bank. As can be seen in Tab.~\ref{tab:size-abla}, the size is not sensitive to generation performance. We obtain the best generation results with the size of 32, which is used for all the experiments in the paper.

\begin{figure}[h]
  \centering
    \includegraphics[width=1\linewidth]{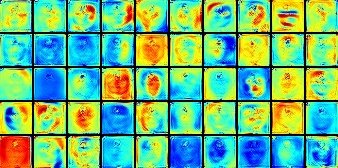}
    \caption{Visualization of randomly selected channels of the meta memory $\mathbf{M}_o$.}
    \label{fig:all_meta}       
\end{figure}

\begin{figure*}[h]
  \centering
    \includegraphics[width=0.60\linewidth]{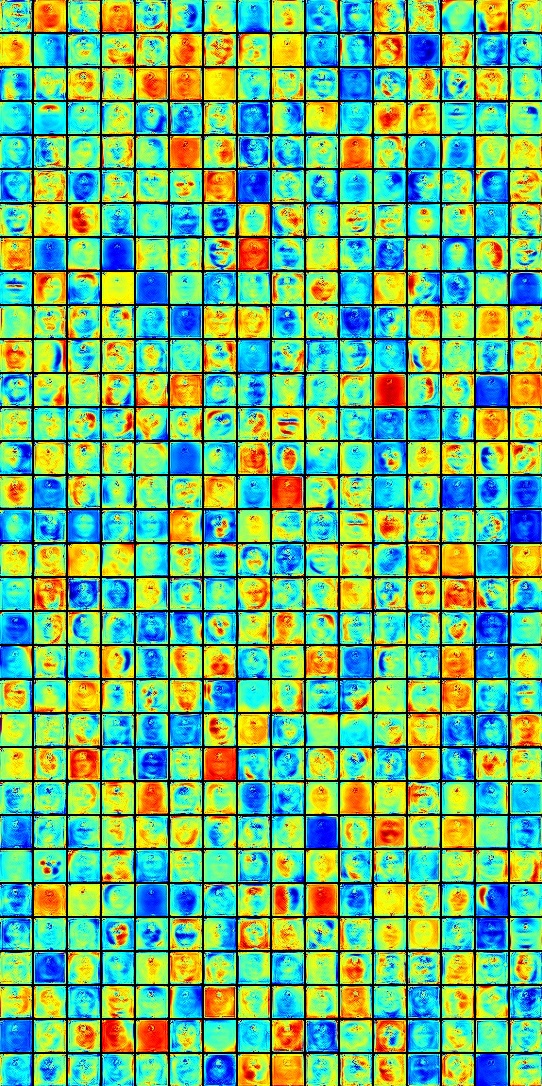}
    \caption{Visualization of all the channels of the meta memory $\mathbf{M}_o$.}
    \label{fig:sub_meta}       
\end{figure*}

\noindent\textbf{Meta memory visualization.} In this section, we show all the channels of our learned meta-memory in Fig.~\ref{fig:all_meta} for better understanding. To better show the details, we also visualize some channels in Fig.~\ref{fig:sub_meta} in high resolution. These visualizations demonstrate the meaningful facial priors are effectively learned in the meta-memory.

\begin{figure*}[h]
  \centering
    \includegraphics[width=0.95\linewidth]{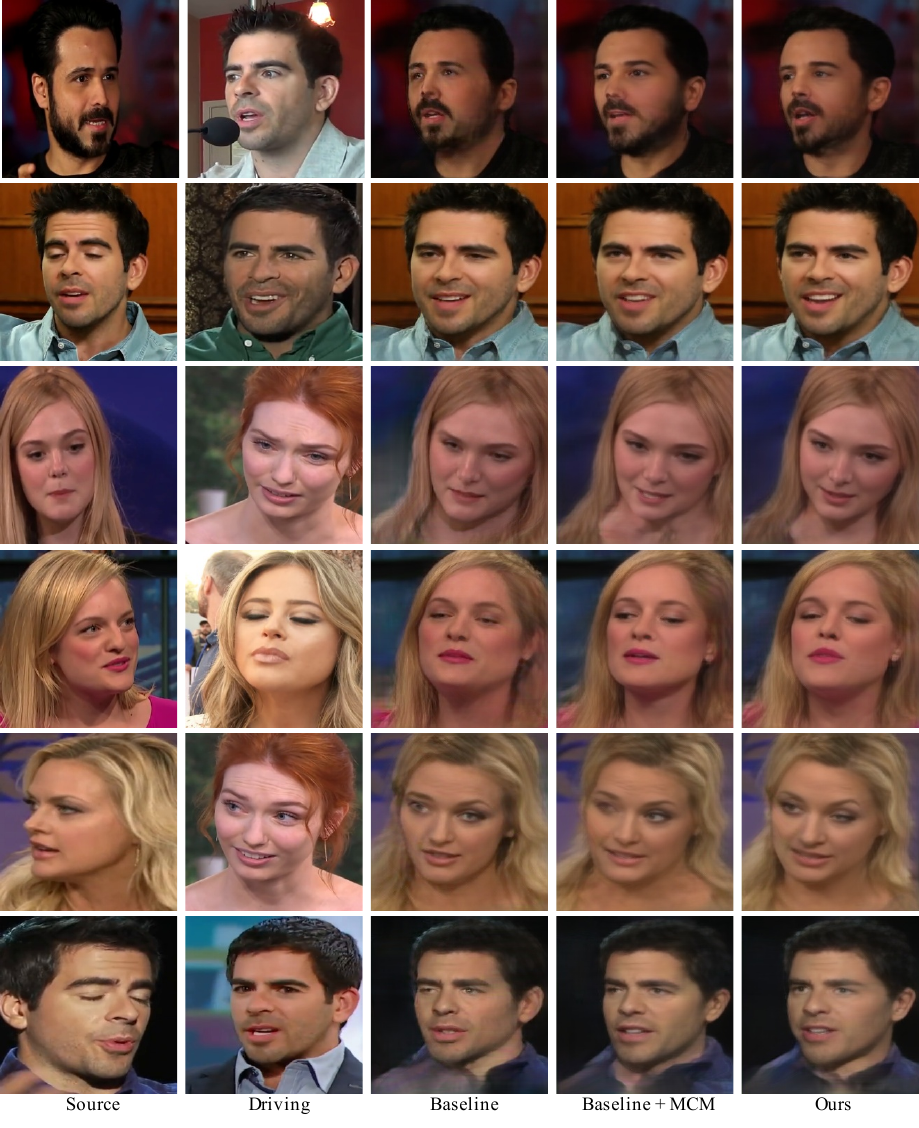}
    \caption{Qualitative ablation studies in VoxCeleb1 dataset. The memory compensation module (MCM) and implicit identity representation conditioned memory module (IICM) can obtain improvements.}
    \label{fig:ablation_study1}       
\end{figure*}

\begin{figure*}[h]
  \centering
    \includegraphics[width=0.95\linewidth]{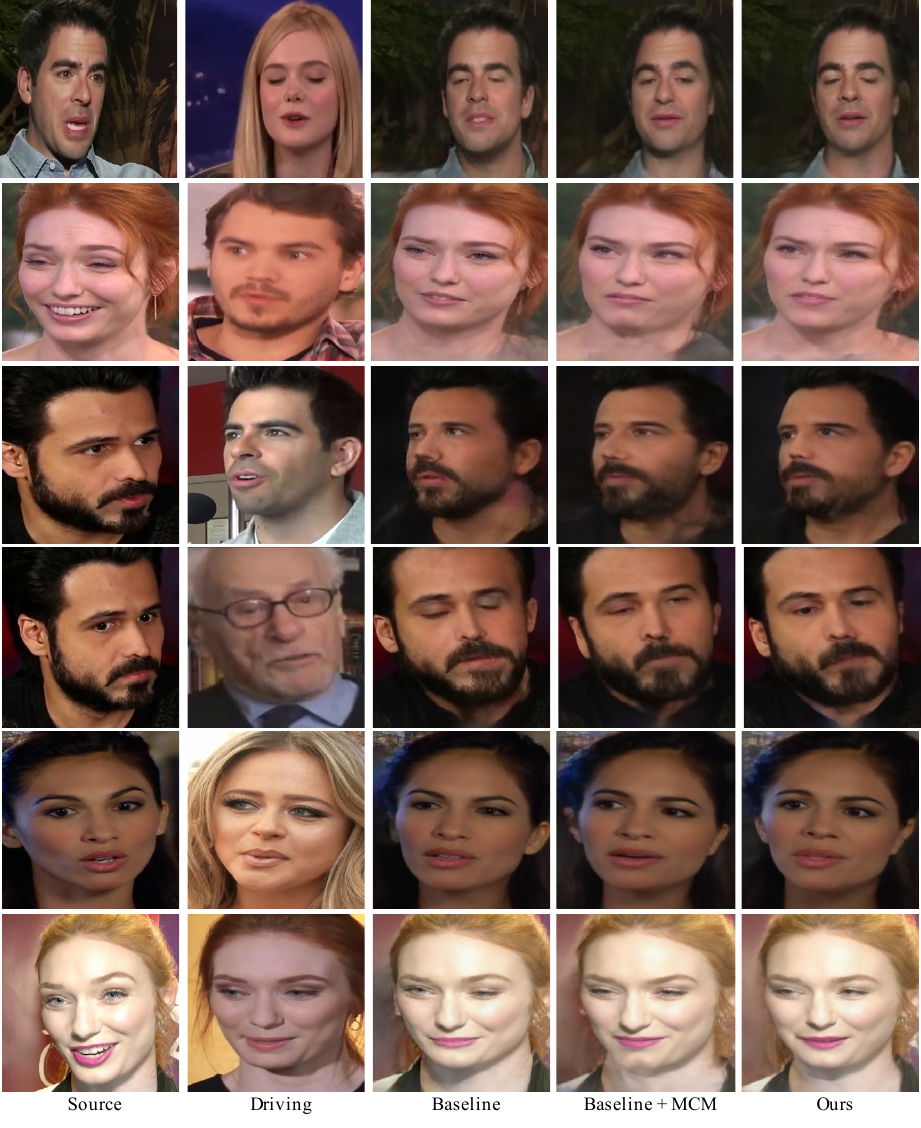}
    \caption{Qualitative ablation studies in VoxCeleb1 dataset. The memory compensation module (MCM) and implicit identity representation conditioned memory module (IICM) can obtain improvements.}
    \label{fig:ablation_study2}       
\end{figure*}

\noindent\textbf{More qualitative ablation studies.} To better show that our designed implicit identity representation conditioned memory compensation network brings improvements, we present more qualitative results for ablation studies in Fig.~\ref{fig:ablation_study1} and Fig.~\ref{fig:ablation_study2}. The effectiveness can be easily observed from the qualitative examples shown in the tables. 

